\documentclass[letterpaper, 10 pt, conference]{ieeeconf}  % Comment this line out if you need a4paper

\IEEEoverridecommandlockouts                              % This command is only needed if 
\usepackage[table,xcdraw]{xcolor}
\usepackage{graphics} % for pdf, bitmapped graphics files
\usepackage{svg}
\usepackage{epsfig} % for postscript graphics files
\usepackage{mathptmx} % assumes new font selection scheme installed
\usepackage{times} % assumes new font selection scheme installed
\usepackage{amsmath} % assumes amsmath package installed
\usepackage{amssymb}  % assumes amsmath package installed
\usepackage{cite}

\usepackage{hyperref}
\hypersetup{%
  breaklinks,
  bookmarksnumbered=true,
  bookmarksopen=true,
  colorlinks=true,
  linkcolor=black,
  urlcolor=black,
  citecolor=black
}

\usepackage{cleveref}
\usepackage{colortbl}
\usepackage{graphicx}
\usepackage{float}
\usepackage{comment}
\usepackage{booktabs}
\usepackage{subcaption}
\usepackage{siunitx}

\title{\LARGE \bf
Allometric Scaling Laws for Bipedal Robots
}

\author{
Naomi Oke$^{1}$, Aja M. Carter$^{1}$, Ben Gu$^{2}$, Cordelia Pride$^{1}$,\\
Steven Man$^{2}$,  Sarah Bergbreiter$^{1}$, and Aaron M. Johnson$^{1,2}$ % <-this % stops a space
\thanks{*This material is based upon work supported by the National Science Foundation under Grants CMMI-2408884, CMMI-2534968, and DGE-2140739. Any opinions, findings, and conclusions or recommendations expressed in this material are those of the author(s) and do not necessarily reflect the views of the National Science Foundation. }% <-this % stops a space
\thanks{$^{1}$Department of Mechanical Engineering, $^{2}$The Robotics Institute, Carnegie Mellon University, Pittsburgh, PA 15213, USA {\tt\small noke@andrew.cmu.edu}}%
}

\def\Zippy{Zippy}
\def\Mugatu{Mugatu}

\begin{document}
\bstctlcite{IEEEexample:BSTcontrol} %This line, along with the text at the top of the .bib file, shortens citations to use et al

\maketitle
\thispagestyle{empty}
\pagestyle{empty}

%%%%%%%%%%%%%%%%%%%%%%%%%%%%%%%%%%%%%%%%%%%%%%%%%%%%%%%%%%%%%%%%%%%%%%%%%%%%%%%%
\begin{abstract}
Legged robots operate across a wide range of physical scales, but how their designs should be adapted as size changes remains unclear. Here, we tackle this question in two ways. First, we survey existing legged robots to provide a broad context for the key scaling variables, robot mass $m$ and leg length $L$. We find the surprising result that bipedal robot mass generally scales with the length squared, $L^2$, rather than the isometric prediction $L^3$. Then, to reduce the variance in design choices, we focus on a pair of previously developed bipeds that share the same quasi-passive morphology but differ by a factor of six in leg length, use different feet and controllers, and achieve different relative speeds. We reconstruct both robots in a common 3-D simulation environment and scale each design over leg lengths from \SIrange{0.02}{1.2}{\meter} under both mass models ($m \propto L^2$ and $\propto L^3$). The controlled comparison shows that velocity follows dynamic similarity, $v\propto L^{1/2}$, across designs and mass models, while the torque needed to sustain walking follows $\tau\propto mL$. Consequently, torque scales approximately with $L^3$ when $m\propto L^2$ and $L^4$ when $m\propto L^3$. A 3-D foot-shape sweep further shows that foot dimensions that permit walking scale approximately linearly with leg length, but the speed-maximizing shape and the mechanism by which each robot moves do not transfer by scaling alone. Overall, the results provide practical insights for rescaling legged systems that leverage natural body dynamics.

\end{abstract}

%%%%%%%%%%%%%%%%%%%%%%%%%%%%%%%%%%%%%%%%%%%%%%%%%%%%%%%%%%%%%%%%%%%%%%%%%%%%%%%%

\section{Introduction}
\label{sec:introduction}

Different tasks and objectives impose different size requirements, creating a need for robots across a wide range of scales. Small robots, for example, can perform micromanufacturing tasks \cite{lassiter2025}, access confined spaces for industrial inspection \cite{caprari2001}, or conduct search and rescue \cite{murphy2004}. Larger robots can address related inspection and exploration tasks with greater payload, mobility, and reach \cite{delmerico2019, frey2026}. When a successful robot design already exists for a task like mobility, adapting that design to a new scale may be preferable to developing an entirely new platform.

Scaling an existing robot is rarely a matter of multiplying every dimension by the same number. Components are not available at arbitrary sizes, actuator capability changes, and a controller that exploits one body's natural dynamics may not transfer to another. Consequently, a geometrically similar robot can be dynamically dissimilar and may require different end effectors, mass distributions, actuators, or gaits.

\begin{figure}[t]
    \centering
    \includegraphics[trim={120 70 150 70},clip,width=\linewidth]{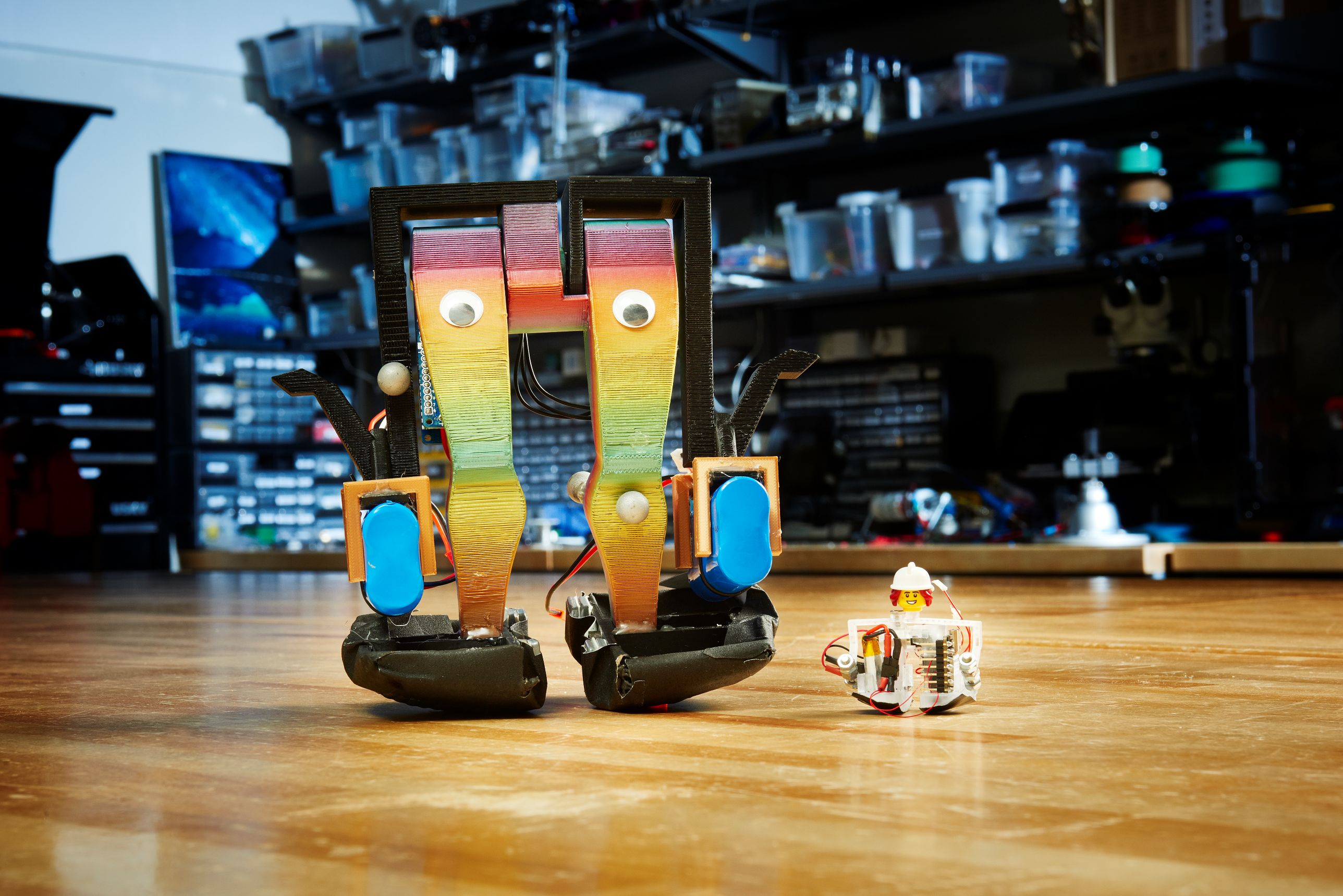}
    \caption{Two robots considered in this paper: \Mugatu{} \cite{mugatu_kyle2023simplest} (left; leg length \SI{15}{\centi\meter}) and \Zippy{} \cite{man-zippy-2025} (right; \SI{2.5}{\centi\meter}).}
    \label{fig:robotpicture}
\end{figure}

These differences appeared directly in the designs of the two robots in \Cref{fig:robotpicture}. \Mugatu{} is a quasi-passive dynamic, hip-actuated biped with a leg length of \SI{0.153}{\meter} \cite{mugatu_kyle2023simplest}. \Zippy{} reduced the leg length to \SI{0.0247}{\meter} while retaining the same two-rigid body and one-actuator in the hip morphology \cite{man-zippy-2025}. The larger robot used sinusoidal position inputs, while the smaller robot could not fit a sensor to achieve that due to its compact size. The smaller robot used ellipsoidal rather than spherical feet, mechanical hard stops, and bang-bang control. \Zippy{} was also faster relative to its size, when compared to the scaled down walking speed of \Mugatu{}. These observations motivate the central question of this paper: \emph{which differences between robot designs such as these are required by scale?}

Isometric and biological scaling provide useful hypotheses for understanding how robot designs might be readapted to operate effectively at different physical scales, but not a complete answer. Under \textit{isometric} scaling, lengths change proportionally and mass follows volume, $m\propto L^3$ \cite{alexander_principles_2003,spletzer1999scaling,mcmahon_muscles_1984}. Dynamic similarity predicts $f\propto L^{-1/2}$ and $v\propto L^{1/2}$ for gravity-dominated locomotion \cite{alexander_principles_2003}. Deviations from these relationships are referred to as \textit{allometric} scaling. Biological systems often depart from isometry because of constant-stress, elastic, metabolic, or evolutionary constraints \cite{mcmahon_size_1973,economos_elastic_1983,ECONOMOS198225,alexander1979allometry}. Robots depart for different reasons: discrete actuators and batteries, manufacturing limits, safety factors, and control architectures do not scale uniformly \cite{burden2024animals}. Biological scaling laws therefore provide candidate relationships to test, not laws that can be transferred from animals without qualification. 
In this paper, we develop several allometric scaling laws for legged robots using a combination of analytical, empirical, and simulation studies.

\subsection{Related Work}
\label{sec:sub_relatedwork}

Isometric scaling with proportional length growth is generally followed in animals, but some key deviations appear. 
Mass generally scales isometrically with body length cubed, $m \propto L_B^{\num{3.00}}$ \cite{alexander1983dynamic} (with other studies ranging from $m \propto L_B^{\num{2.80}}$ \cite{silva_allometric_1998} to $m \propto L_B^{\num{3.18}}$ \cite{economos_elastic_1983})\footnote{
Note that in the biology literature, scaling is generally done with respect to mass (i.e.\ $L \propto m^{1/3}$), however in terms of robot design using length as the base is more applicable.}. 
However, mass grows a little slower than the cube of limb length, or equivalently limbs get relatively longer as the animal gets larger.
For example, one study found that mammalian mass versus bone length scales as $m \propto L^{\num{2.85}}$ \cite{alexander1979allometry}, while another found $m \propto L^{\num{2.5}}$ for forelimbs and $m \propto L^{\num{2.7}}$ for hindlimbs of quadrupedal mammals \cite{kilbourne2013scale}. This allometry is even more evident in avian bipeds, which exhibit $m \propto L^{\num{2.44}}$ \cite{daley2018scaling}, suggesting that bipedalism may require different scaling than other forms of locomotion. 

Beyond morphology, locomotor speed also exhibits scale dependence. Past work demonstrated that maximum speed for reptiles, birds, and mammals follows a hump-shaped scaling law, explaining why the largest animals are not the fastest, due to acceleration limits and metabolic constraints \cite{hirt_general_2017,labonte2024dynamic}. Before that peak, velocity scales as $v \propto L^{\num{0.25}}$ \cite{labonte2024dynamic}. 

Recent cross-domain analyses reinforce an ambiguity in torque and power in biology and robotics. Under isometric similarity, rotational arguments predict joint torque demands on the order of $\tau \propto mL^{2} \propto L^{3} \cdot L^{2}= L^{5}$. However, both biological and robotic systems routinely deviate from this upper-bound scaling. Models of torque profiles have been used to understand scale-dependent active muscle responses in biological hip joints, showing that torque demand appears to scale as only $\tau \propto L^3$ \cite{young_analyzing_2022}. Some analyses extend beyond biological muscle power, body mass, and acceleration time to a general power-mass-speed relationship. Studies of animals and ground-mobile machines find that both animals and ground-mobile machines obey a common power-mass-speed scaling law, suggesting deep physical constraints that transcend implementation \cite{kott_cockroaches_2021,heglund_energetics_1982}.

In robotics, however, scaling laws are less well established and are not limited by evolutionary, ecological, and biological constraints. Isometric scaling, derived from geometry and first principles, provides a natural starting point for scaling robotic systems \cite{spletzer1999scaling}. It is therefore widely used in robotics, but it does not account for allometric effects that arise when different properties scale at different rates, such as component mass, stiffness, actuator performance, and manufacturing constraints. Consequently, purely isometric scaling often fails to predict how real systems built at different scales will behave. For example, \cite{jayaram2020scaling} found that elastic similarity more accurately described the scaling of their insect-scale quadrupedal robots than geometric similarity.

\begin{figure*}[t!]
\centerline{\includegraphics[width=0.88\linewidth]{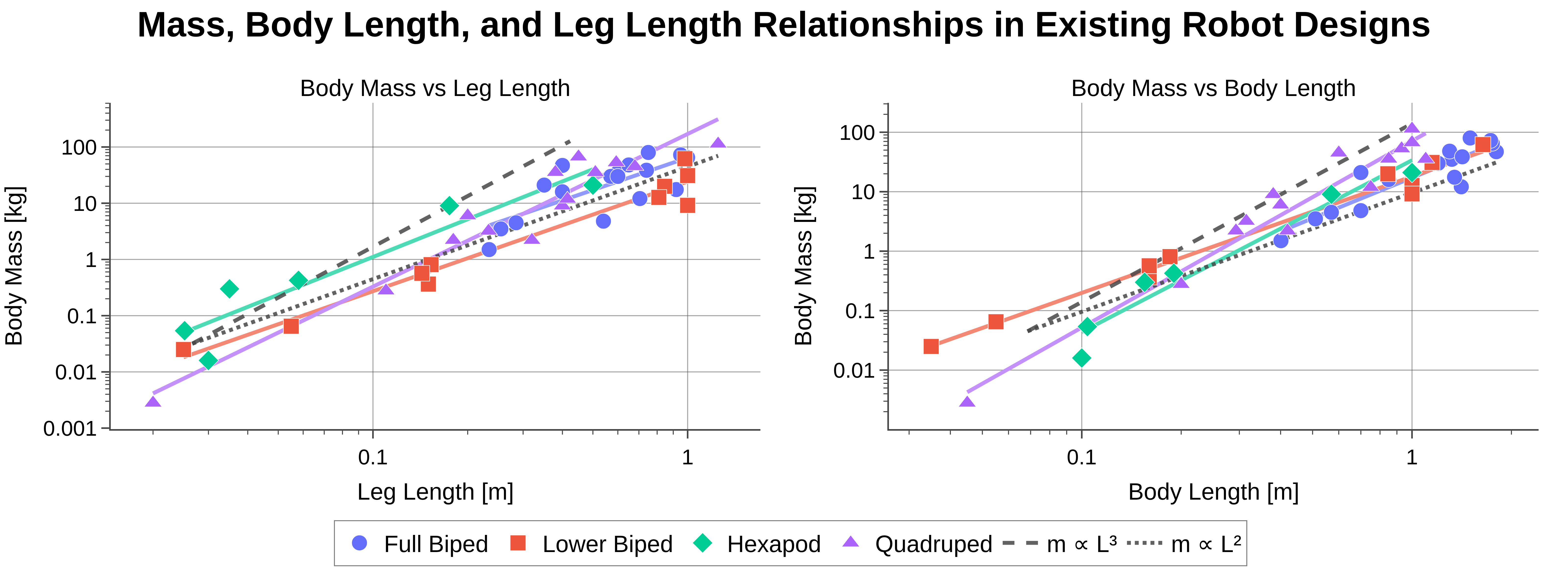}}
\caption{Relationships between body length, leg length, and body mass across multiple existing robot morphologies.}
\label{fig:robotreview}
\end{figure*}

\begin{table*}[t!]
    \centering
    \caption{Empirical evidence for robotic allometry from a survey of existing robot designs.}
    \begin{tabular}{rlllllll}
    \toprule
         & \textbf{Lower Bipedal Robot}&\textbf{Full Bipedal Robot}& \textbf{Quadrupedal Robot} & \textbf{Hexapedal Robot} & \textbf{Global}\\
         \midrule
        Number in group $n=$ & \num{10} & \num{20} & \num{14} & \num{6} & \num{50}\\
         % Number in group $n=$ & \textcolor{black}{\num{10}} & \textcolor{black}{\num{17}} & \textcolor{black}{\num{14}} & \textcolor{black}{\num{6}} & \textcolor{black}{\num{47}}\\
         Body Length [\si{\meter}] $L_B \propto $& $L^{\num{1.00}}$ ($R^2$=\num{0.99}) & $ L^{\num{0.86}}$ ($R^2$=\num{0.75}) & $L^{\num{0.81}}$ ($R^2$=\num{0.87}) & $L^{\num{0.79}}$ ($R^2$=\num{0.98}) & $ L^{\num{0.87}}$ ($R^2$=\num{0.87})\\
         Body Mass [\si{\kilo\gram}] $m \propto$ &  $L^{\num{1.95}}$ ($R^2$=\num{0.96}) &  $ L^{\num{1.91}}$ ($R^2$=\num{0.59}) & $ L^{\num{2.72}}$ ($R^2$=\num{0.93})& $ L^{\num{2.23}}$ ($R^2$=\num{0.88})& $ L^{\num{2.12}}$ ($R^2$=\num{0.87})\\
         Body Mass [\si{\kilo\gram}]  $m \propto$  &  $L_B^{\num{1.96}}$ ($R^2$=\num{0.98})& $ L_B^{\num{2.27}}$ ($R^2$=\num{0.83}) & $ L_B^{\num{3.14}}$ ($R^2$=\num{0.94})& $ L_B^{\num{2.89}}$ ($R^2$=\num{0.95})& $ L_B^{\num{2.32}}$ ($R^2$=\num{0.91})\\
         \bottomrule
    \end{tabular}
\label{tab:robotscaling}
\vspace{-1em}
\end{table*}

As for how motors and required torque scale, there is a disagreement in the literature \cite{rodriguez2026}. Theoretical models for torque relative to motor diameter range from $\tau \propto L^{\num{3.5}}$ \cite{caprari2001,haddadin2012rigid}, to $\tau \propto L^4$ \cite{honsinger1987sizing,chatzakos2008influence,reichert2009torque}, to as high as $\tau \propto L^5$ \cite{spletzer1999scaling,caprari2001}, depending on assumptions about magnetic loading, current density, and cooling.
%Several papers also consider how torque scales with various fixed parameter \cite{seok2012Actuator,seok2015design,kenneally2016Design}, which all extend to $\tau \propto L^4$ if full scaling is considered.
Empirical studies span a similar range, including $\tau \propto L^3$ \cite{dermitzakis_scaling_2011}, $\tau \propto L^{\num{3.5}} $\cite{waldron2000scaling,haddadin2012rigid},  $\tau \propto L^{\num{3.8}}$ \cite{dermitzakis_scaling_2011,seok2015design,sarans_parallel_motor-gearbox_2020}, $\tau \propto L^4$ \cite{chatzakos2008influence,kenneally2016Design}, and $\tau \propto L^5$ \cite{spletzer1999scaling}. 
Beyond torque, the stress-limited speed appears to scale as $\omega \propto L^{-1}$, constraining the joint power range at large sizes \cite{spletzer1999scaling, caprari2001, chatzakos2008influence,waldron2000scaling}. The scaling of the gearbox and transmission also plays an important role \cite{saerens2019scaling}.

More recently, Miller and Clark \cite{miller_clark_IROS2015} developed a generalized framework for dynamically scaling legged robots using force balance and Buckingham-$\pi$ dimensional analysis, revealing a theoretical torque relationship. Rather than assuming constant density ($m \propto L^3$), their formulation preserves dynamic similarity while allowing mass to scale independently through a free force scale factor. Under gravity, this yields the dimensional relationship, $T \propto L^{1/2}$, implying that stride frequency scales as $f \propto L^{-1/2}$, walking velocity scales as $v \propto L^{1/2}$, and joint torque scales as $\tau \propto mL$, rather than the isometric prediction $\tau \propto mL^2$. They validated these relationships on planar reduced-order locomotion models and plausible redesign of the hexapedal robot series, RHex.

\subsection{Contribution}

The \cite{miller_clark_IROS2015} framework informs robot scaling to preserve dynamic similarity and allows for further research into how to design robots when dynamic similarity is not followed exactly. In particular, real redesigns often violate an exactly scaled dynamic similarity transformation because control architecture, contact geometry, available components, and mass distribution change, making it important to define mass relationships before ascribing a scaling similarity assumption. Here, we study what remains predictable under those departures by instead asking what mass scaling is observed among existing legged robots and using those data to select a practical mass prior for scaled legged robot design. 

To address that gap and understand scaling for legged systems, we first survey existing robots to estimate how mass and body proportions change in practice. This survey expands to many legged morphologies and supplies an empirical mass scaling model (\Cref{sec:existingrobots}). Second, we place two specific robots, \Cref{fig:robotpicture}, in the same simulation environment and scale each independently, allowing scale effects to be separated from differences in foot shape and control (\Cref{sec:simmethod}). We use these systems because of their limited control and actuation complexity, allowing for better isolation of relationships between design parameters and an understanding of natural body dynamics for bipedal walking \cite{mcgeer1990passive}. Third, we identify the relationships that transfer across both designs and the gait mechanisms that remain design-specific, particularly velocity, torque demand, and the boundary of feasible foot shapes that lead to walking, as quasi-passive bipedal walkers are heavily influenced by foot shape (\Cref{sec:simresults}) 
%\textbf{(There are a lot of citations that all kinda say the same thing so please help me pick one)} \cite{smyrli2020, smyrli2019, kwan2007, fukushima2014, hyodo2008, martin2012}
\cite{kwan2007}. Finally, we turn these findings into a redesign workflow for scaling robots and discuss what would be needed to extend this study to other robots (\Cref{sec:discussion}). 

\section{Survey of Existing Robot Designs} 
\label{sec:existingrobots}

%Before scaling either case-study robot, we require a plausible relationship between leg length and mass. 
We begin by looking across designs to observe the general trends between lengths and masses on real, existing robots, and compare these to the isometric relationship, $m\propto L^3$, though robots do not necessarily need to preserve constant density. To test this, we conduct a literature survey to estimate how existing legged robots scale in practice. The survey spans bipedal, quadrupedal, and hexapedal platforms and records leg length, overall body length, mass, nominal speed, and degrees of freedom (DOF). Bipedal robots were split into two groups based on the morphology and degrees of freedom. Bipedal robots designed without an upper torso or with less than \num{4} DOFs were sorted into the ``lower bipedal'' category. Other bipedal robots were classified as ``full bipedal'' robots. 

In total, we included data for \num{10} lower bipeds, \num{20} bipeds, \num{14} quadrupeds, and \num{6} hexapods. 
% The full data table, including sources, is in an \href{https://github.com/robomechanics/robot-dataset}{open source repository}. 
The full data table and sources will be released in an open-source repository. The analyzed robots include full bipeds Atlas HD, Digit, Unitree G1, 
% Unitree H1, Unitree H2, Unitree R1, Unitree R1 Air, 1X NEO, 
and Booster T1; 
% Asimo, Digit, NAO, Berkeley Humanoid Lite, Robotis OP3, CHARLI, ARTEMIS, MIT humanoid, Apollo, Kondo KHR-3, NimbRo-OP2, Xuan Feng Xiao Zi N2, BRUCE, Squeaky, Mugatu, Zippy, Limbx Tron1, 
lower bipeds include Cornell Ranger, Cassie, Mugatu, and Zippy; 
%ATRIAS, Ranger, 
quadrupeds include 
%BigDog, Cheetah 3, Spirit,Minitaur,Vision,
Mini Cheetah, Spot, Petoi Bittle, and HAMR-F;
%ANYmal, Solo, Loris, Aibo, Llama,  SiLVIA,
and hexapods include X-RHex, MiniRHex, DASH, and iSprawl.
%X2-VelociRoACH, M4, Daisy, CMU Tappy, and CMU Tappina. 
The regression results are summarized in \Cref{tab:robotscaling} and \Cref{fig:robotreview}. 

Leg length, $L$, is defined as the robot's hip height in a nominal standing posture. Body length is defined as the robot's longest dimension in a nominal standing posture, generally height for bipeds and snout-ventral length\footnote{Here, body length is defined using the biological convention of snout-ventral length, from the tip of the nose to base of the tail} for quadrupeds and hexapods. We find that lower bipedal robots, which are primarily legs, followed the isometric relationship of $L_{B} \propto L^{\num{1.0}}$. In contrast, we observe that for all of the other robots, \textbf{the body length $L_{B}$ scales more slowly than the leg length $L$, approximately $L_B\propto$ with exponents of \num{0.86} or less}, showing a divergence from the relationships assumed from isometric scaling ($ L_B \propto L^1$). 
On average, as robots get bigger, their legs become relatively larger compared to their body. This trend is similar to that observed in biology \cite{alexander1979allometry,kilbourne2013scale,daley2018scaling}, suggesting that this is a property of locomotion behavior.

%(Here, body length is defined snout-ventral length (from the tip of the nose to base of the tail) and in amphibians and reptiles, and $M^{-1/3}$ as proxy for size in most other animal studies. 
%Generally, observed leg length follows positive allometry.
%We find that these robots have a relationship between body length and leg length of about $L_{B} \propto L^{\textcolor{black}{\num{0.87}}}$.

Similarly, mass does not follow one exponent across every morphology. For both biped groups, \textbf{mass surprisingly scales with leg length at approximately $m\propto L^2$}, below the isometric prediction $m\propto L^3$. 
This difference may be the result of several factors: mass for computation does not need to scale as fast, larger motors are generally better than small motors, and manufacturing constraints make fabricating small robots harder than larger robots \cite{rodriguez2026}.
The survey also reveals more variability in the full biped group than any other group, showing a wide range of achievable designs (\Cref{tab:robotscaling}). Quadrupeds and hexapods scale more steeply with leg length and approach $m\propto L_B^3$ when mass is expressed against body length. 

While these regressions only describe the robots represented in the dataset, they provide a starting point for future designs and analyses. Other metrics like speed, efficiency, or mass distribution are not reported consistently enough in the literature to result in statistically significant trends. 

%The two key allometric scaling laws from this survey are that body length grows slower than leg length in all but the lower biped categories, and that mass grows more slowly than $L^3$, particularly for bipedal robots. 

%For the case study, however, they justify testing both $m\propto L^2$, as the empirically motivated biped model, and $m\propto L^3$, as the constant-density baseline.

\section{Controlled Two-Robot Scaling Study} 
\label{sec:simmethod}

\begin{figure}[t]
\centering
\includegraphics[width=0.8\columnwidth]{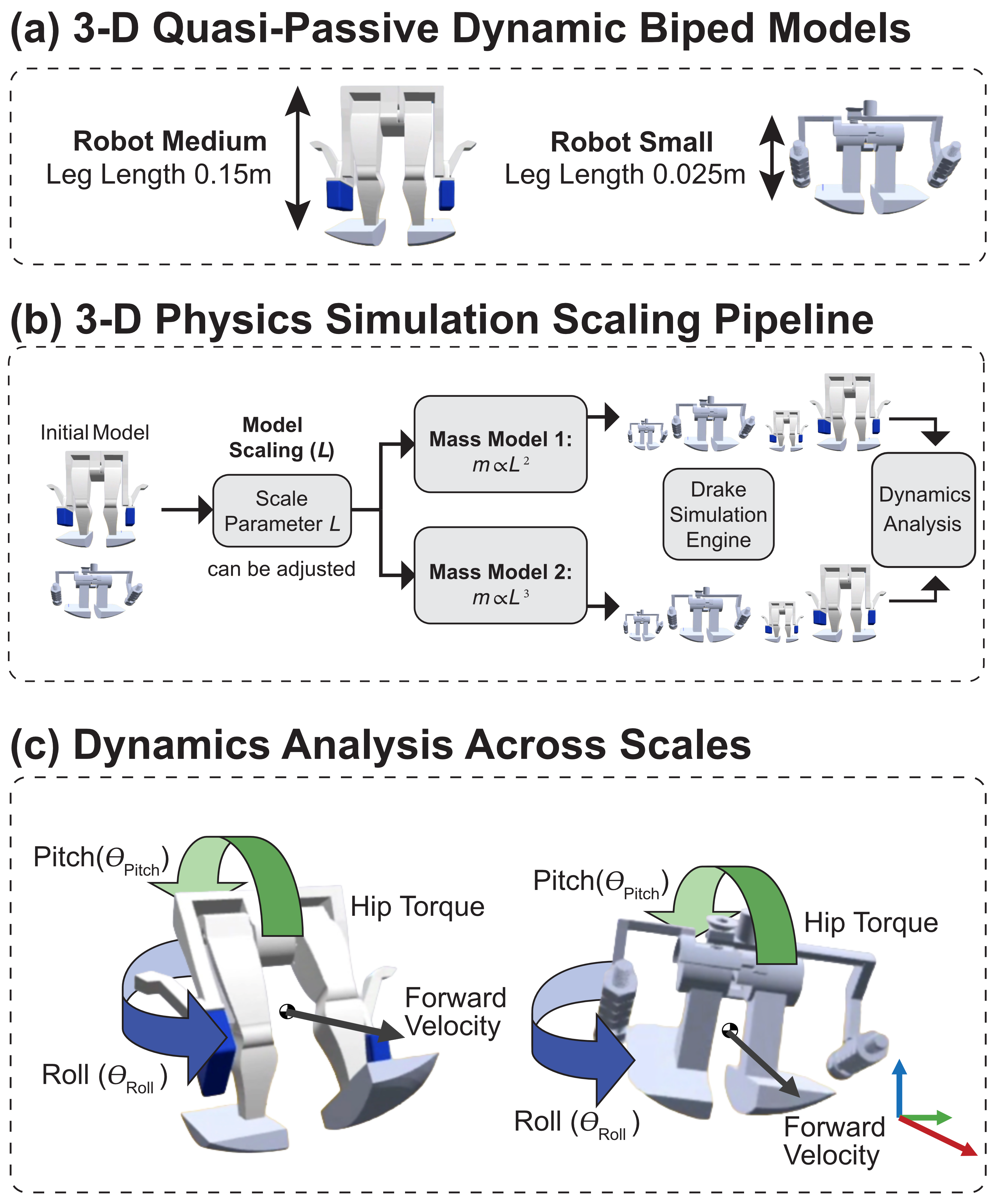}
\caption{(a) The two simulated robot models have nominal leg lengths of \SI{0.153}{\meter} and \SI{0.0247}{\meter}. (b) Each design is geometrically scaled by leg length $L$, with mass following either $m\propto L^2$ or $m\propto L^3$, and evaluated in the Drake simulation environment. (c) Forward velocity, hip torque, and body attitude are recorded across scales to identify shared scaling trends and design-specific locomotion dynamics.}
\label{fig:mugatu_zippy_sim}
\end{figure}

\begin{table}[t]
\centering
\caption{Physical parameters of robot configurations tested in simulation.}
{
\begin{tabular}{ l c c}
\toprule
 & \textbf{\Mugatu{}} 
% & \textbf{Mugatu-Zippy Hybrid} 
& \textbf{\Zippy{}} \\ 
\midrule
Foot Shape & Spherical & Ellipsoidal\\
Leg Length  [\si{\milli\meter}] $L$   & \num{153.0}    & \num{24.7}    \\ 
Body Length  [\si{\milli\meter}] $L_B$ & \num{186.0}    & \num{36.4}    \\
$Y, Z$ Semi-Axis [\si{\milli\meter}] $R_Y,R_Z$ & \num{120.0}     & \num{25.0} \\ 
$X$ Semi-Axis  [\si{\milli\meter}] $R_X$ & \num{120.0}    & \num{30.0}  \\ 
Body Mass  [\si{\gram}] $m$   & \num{900.0}    & \num{25.0}    \\ 
\bottomrule
% \multicolumn{7}{l}{\small \textit{Estimation from CAD}}
%\multicolumn{3}{l}{\small \textit{Estimated from CAD}}
\end{tabular}
}
\label{tab:robot_dims}
\end{table}

While the survey revealed some interesting allometric scaling laws, the variations in designs and inconsistent data reporting limited its utility to only basic trends. To provide a more direct one-to-one comparison with consistent methodology, here we present a case study comparing \Mugatu{} and \Zippy{} (Figures~\ref{fig:robotpicture} and~\ref{fig:mugatu_zippy_sim}), and in particular why they required different feet and achieved different scaled forward velocities. We therefore scale both robots in a common simulation while preserving the defining morphology and controller of each. This paired design enables us to identify trends that persist across robot designs while distinguishing them from mechanisms specific to an individual walker, all within a set of sufficiently similar morphologies.

We simulated a series of bipedal robots in a 3-D simulator, Drake \cite{tedrake2019drake}. This simulator was chosen because of its hydroelastic contact model, which allows high-fidelity contact modeling for systems like passive dynamic walkers that have rolling contact. To ensure that the simulator was able to capture the behaviors of the real hardware, simulations at the \num{1}$\times$ scale were compared to experimental results on hardware. The simulator was set up to use the same mass distribution, body shape, and actuation parameters (i.e.\ sinusoidal control vs bang-bang control, motor torque-speed relationship) as the real hardware. The hydroelastic contact parameters were tuned until the body roll, pitch, and yaw matched those achieved by the real hardware. Achieving reliable results in simulation is difficult, particularly at small scales, due to friction, inertial effects, and contact forces. To ensure matching at smaller scales, the mesh resolution parameter was decreased proportionally in accordance with Drake guidelines \cite{tedrake2019drake}.

% The full simulation is available in an \href{https://github.com/robomechanics/bipedal-scaling-laws}{open source repository}.
The full simulation will be released in an open-source repository. Though this study is exclusively for quasi-passive dynamic walkers, their dynamics are governed by pendular dynamics and gravitational loading, which is common across all designs utilizing natural dynamics.

CAD models of each robot were obtained from the authors of \cite{mugatu_kyle2023simplest,man-zippy-2025} (\Cref{fig:mugatu_zippy_sim} and \Cref{tab:robot_dims}). Each variation was simulated at a range of scales between $L = \SIrange{0.02}{1.2}{\meter}$, including points at the leg lengths of \Zippy{} (\SI{0.0247}{\meter}), \Mugatu{} (\SI{0.153}{\meter}), and a large human size (\SI{1.0}{\meter}).  
When scaling these designs in simulation, all lengths were scaled equally, matching the relationship found for lower bipedal robots in the previous section. 
Mass was scaled with both mass scaling models ($m \propto L^2$ and $m \propto L^3$), and the minimum torque required to achieve consistent walking was manually tuned. Stride frequency was scaled to maintain the pendular dynamics that form the basis of passive-dynamic walking. 

During single support, each robot behaves as a physical pendulum pivoting about the stance foot. Its natural frequency is $\omega = \sqrt{\frac{m g d}{I}}$, where $d$ is the distance from the stance pivot to the center of mass and $I$ is the moment of inertia about that pivot. If the mass distribution remains geometrically similar, which holds by construction in our simulations, then $d \propto L$ and $I \propto mL^2$. Substituting:
\begin{equation}
\omega = \sqrt{\frac{m g L}{m L^2}} = \sqrt{\frac{g}{L}} \propto L^{-1/2}
\label{eq:pendular_freq}
\end{equation}

The body mass $m$ cancels identically. Stride frequency therefore scales as $f \propto \omega \propto L^{-1/2}$, while step length is set by the geometry of the leg swinging. From this, we expect walking velocity to be $v = f \cdot (\text{step length}) \propto L^{-1/2}\cdot L = L^{1/2}$, recovering the Froude relationship and a scale-invariant Froude number Fr $= \frac{v^2}{gL}\propto 1$ from dynamic similarity \cite{alexander1983dynamic}.

During each simulation, body attitude, body position, and other body-state data are recorded. For each condition, the robot's locomotion is evaluated after stable, consistent walking was achieved, defined as the time at which the body pitch and pitch rate converged to a 2-D limit cycle. We record net forward center-of-mass progress per unit time over a fixed window, normalized by leg length ($v_f/L$) for cross-size comparison. These data are then analyzed to identify trends in body dynamics and locomotion.

To determine the amplitudes of body attitude oscillation, roll, pitch, and yaw were measured for the duration of the walking trial, and the peaks and valleys of the signal were used to determine the average half-peak amplitudes. 

Robot design is further evaluated through a controlled study of foot-shape scaling to explore its effect on body dynamics. Both robots are compared at the three characteristic body scales with each robot's foot curvature independently scaled along the X, Y, and Z axes from $\num{0.5}\times$ to $\num{2.5}\times$ in increments of $\num{0.1}$. For every resulting foot-shape variation, the robot is simulated for \SI{30.0}{\second}. Within a given robot and leg length, total mass and commanded frequency are held constant across all foot shapes.

To analyze foot-shape scaling, we also record the retraction fraction $\rho$, defined as the backward distance traveled by the leg during swing relative to the robot's forward distance traveled per stride. A robot advancing steadily without ever moving backward has $\rho = \num{0}$. A robot rocking in place covers equal backward and forward path and has $\rho = \num{0.5}$, and one drifting backward has $\rho > \num{0.5}$. We use this value to sort and analyze the data from the foot-shape scaling. From preliminary analysis, we find that \SI{97.23}{\percent} of non-walking trials fall between $\num{0.40}$ and $\num{0.60}$ with median $\num{0.47}$, and \SI{23.42}{\percent} exceed $\num{0.50}$, having traveled net backward. We therefore define walked trials as $\rho = \num{0}$ to $\num{0.43}$. For each foot-shape trial, two independent criteria are available: \emph{walked}, meaning $\rho \leq \num{0.43}$, and \emph{upright}, marking whether the robot toppled during the trial. Correlations for the foot-shape scaling analysis are Pearson correlations, $r$, and are computed within a single robot and leg length.

%while Retraction and path efficiency, $\eta$ satisfy $\eta = 1 - 2\rho$, so $\rho$ is a rescaling of $\eta$ rather than an independent measurement. We define \emph{walked} as $\eta \geq \textcolor{black}{\num{0.15}}$.

\section{Results: Understanding Predictive Scaled Relationships} 
\label{sec:simresults}

\subsection{Mass, Frequency, and Velocity Scaling}
\label{sec:sub_mfv}

\begin{figure}[t]
\centering
\scriptsize
\includegraphics[width=0.91\columnwidth]{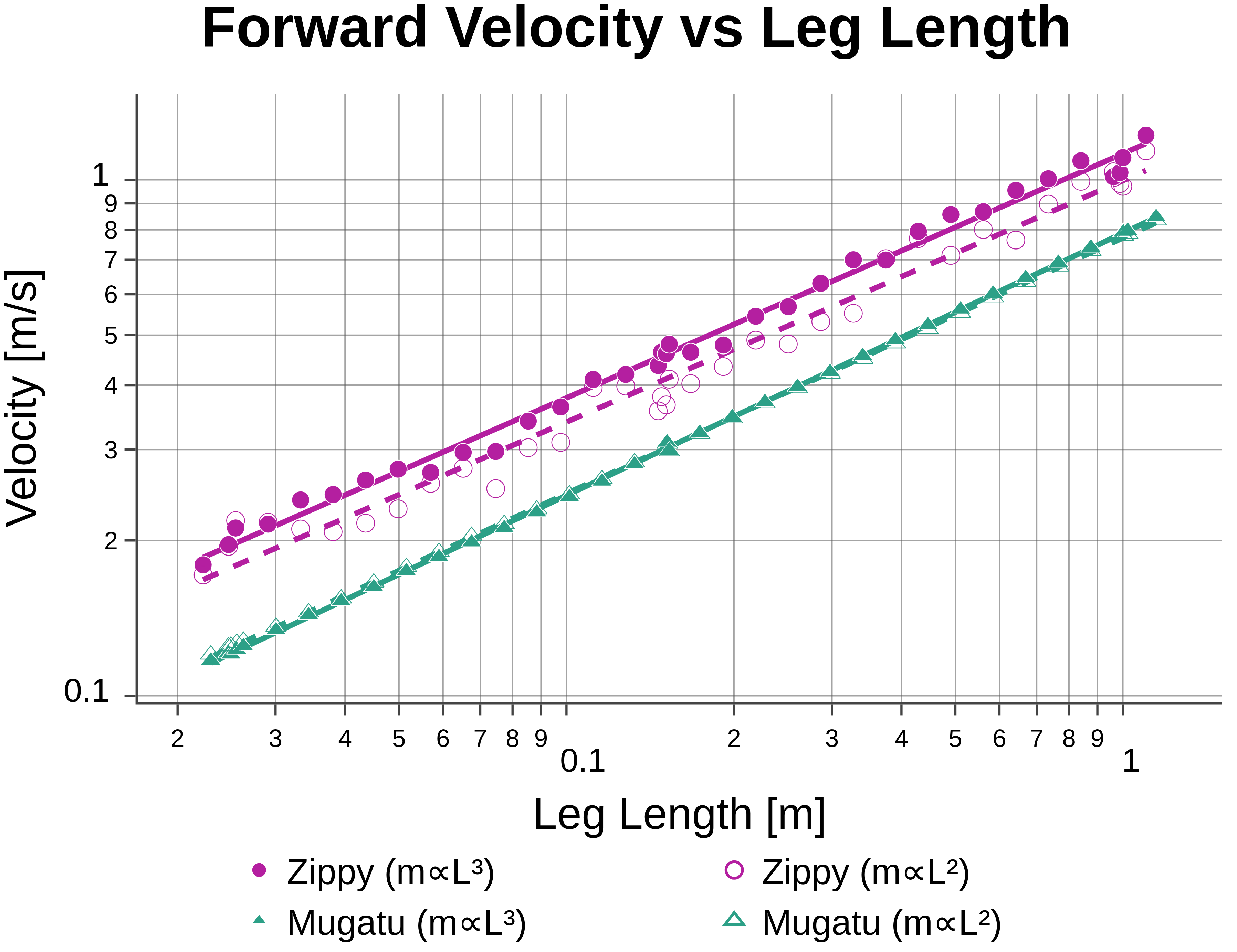}
%\vspace{-2em}
\caption{Simulated forward velocity of \Zippy{} and \Mugatu{} across scales (see \Cref{tab:simresults}).}
\label{fig:velocity}
\end{figure}

Across both robot designs, both mass models, and foot shapes, \textbf{walking velocity increases with leg length by approximately $v_f \propto L^{\num{0.5}}$} (\Cref{fig:velocity,tab:simresults}), matching the dynamic-similarity prediction. 
%Changing the mass exponent from two to three has little effect on the velocity exponent. 
Within this case study, leg length and gait time scale determine how forward velocity changes with size, whereas mass magnitude does not. Note also that leg-length-normalized velocity decreases with size under both mass laws for both robots (\Cref{tab:simresults}). Larger versions are therefore faster in absolute terms but slower in body lengths per second.

\begin{table}[t]
\centering
\caption{Simulation results for velocity, torque, body attitude, and foot-shape scaling. \emph{Forward velocity $v_f$} is the nominal-design speed and \emph{Normalized $v_f/L$} is forward velocity normalized by leg length; \emph{Optimized $v_O$} is the maximum velocity over the foot-shape sweep for the 3 leg scales. 
%The foot-optimized peak at $L=\SI{1.0}{\m}$ is within a few percent of Froude extrapolation from each robot's nominal scale (\Zippy{} \SI{+11}{\percent}, \Mugatu{} \SI{-2}{\percent} under $m\propto L^3$; \Zippy{} \SI{-3}{\percent}, \Mugatu{} \SI{-2}{\percent} under $m\propto L^2$).
}
\label{tab:simresults}
\setlength{\tabcolsep}{2.5pt}
\renewcommand{\arraystretch}{1.05}
\resizebox{\columnwidth}{!}{%
\begin{tabular}{@{}lcc@{}}
\toprule
 & \textbf{\Zippy{}} & \textbf{\Mugatu{}} \\
\midrule
\multicolumn{3}{c}{Locomotion Performance for $m \propto L^2$ Scaling} \\
\midrule
Forward $v_f$ [\si{\meter\per\second}]
 & $L^{\num{0.47}}$ ($R^2=\num{0.98}$)
 & $L^{\num{0.50}}$ ($R^2=\num{1.00}$) \\
Normalized $v_f/L$ [\si{\per\second}]
 & $L^{\num{-0.53}}$ ($R^2=\num{0.98}$)
 & $L^{\num{-0.50}}$ ($R^2=\num{1.00}$) \\
Optimized $v_O$ [\si{\meter\per\second}]
 & $L^{\num{0.49}}$
 & $L^{\num{0.51}}$ \\
Minimum $\tau$ [\si{\newton\meter}]
 & $L^{\num{2.92}}$ ($R^2=\num{1.00}$)
 & $L^{\num{3.49}}$ ($R^2=\num{0.97}$) \\
$\theta_R$ [\si{\degree}]
 & $\num{9.28}\pm\num{0.59}$
 & $\num{10.83}\pm\num{0.13}$ \\ 
$\theta_P$ [\si{\degree}]
 & $\num{25.02}\pm\num{3.05}$
 & $\num{19.91}\pm\num{0.32}$ \\ 
$\theta_Y$ [\si{\degree}]
 & $\num{7.13}\pm\num{1.12}$
 & $\num{10.1}\pm\num{0.3}$ \\
\midrule
\multicolumn{3}{c}{Locomotion Performance for $m \propto L^3$ Scaling} \\
\midrule
Forward $v_f$ [\si{\meter\per\second}]
 & $L^{\num{0.47}}$ ($R^2=\num{0.99}$)
 & $L^{\num{0.51}}$ ($R^2=\num{1.00}$) \\
Normalized $v_f/L$ [\si{\per\second}]
 & $L^{\num{-0.53}}$ ($R^2=\num{0.99}$)
 & $L^{\num{-0.49}}$ ($R^2=\num{1.00}$) \\
Optimized $v_O$ [\si{\meter\per\second}]
 & $L^{\num{0.53}}$ 
 & $L^{\num{0.52}}$ \\
Minimum $\tau$ [\si{\newton\meter}]
 & $L^{\num{3.90}}$ ($R^2=\num{1.00}$)
 & $L^{\num{4.07}}$ ($R^2=\num{0.98}$) \\
$\theta_R$ [\si{\degree}]
 & $\num{10.09}\pm\num{1.23}$ 
 & $\num{10.82}\pm\num{0.24}$ \\ 
$\theta_P$ [\si{\degree}]
 & $\num{27.69}\pm\num{3.06}$
 & $\num{19.91}\pm\num{0.32}$ \\ 
$\theta_Y$ [\si{\degree}]
 & $\num{8.43}\pm\num{2.21}$
 & $\num{9.82}\pm\num{0.39}$ \\
\midrule
\multicolumn{3}{c}{Feasible foot shapes for $m \propto L^3$ Scaling} \\
\midrule
$R_X$ [\si{\meter}]
 & $\num{1.80}L^{\num{0.95}}$ ($R^2=\num{0.96}$)
 & $\num{1.43}L^{\num{0.99}}$ ($R^2=\num{0.97}$) \\
$R_Y$ [\si{\meter}]
 & $\num{1.50}L^{\num{0.98}}$ ($R^2=\num{0.94}$)
 & $\num{1.27}L^{\num{0.99}}$ ($R^2=\num{0.95}$) \\
$R_Z$ [\si{\meter}]
 & $\num{1.23}L^{\num{1.00}}$ ($R^2=\num{0.90}$)
 & $\num{0.92}L^{\num{1.03}}$ ($R^2=\num{0.91}$) \\
\midrule
\multicolumn{3}{c}{Feasible foot shapes for $m \propto L^2$ Scaling} \\
\midrule
$R_X$ [\si{\meter}]
 & $\num{2.03}L^{\num{0.99}}$ ($R^2=\num{0.97}$)
 & $\num{1.42}L^{\num{0.99}}$ ($R^2=\num{0.97}$) \\
$R_Y$ [\si{\meter}]
 & $\num{1.60}L^{\num{1.00}}$ ($R^2=\num{0.95}$)
 & $\num{1.30}L^{\num{1.01}}$ ($R^2=\num{0.96}$) \\
$R_Z$ [\si{\meter}]
 & $\num{1.32}L^{\num{1.03}}$ ($R^2=\num{0.92}$)
 & $\num{0.95}L^{\num{1.00}}$ ($R^2=\num{0.91}$) \\
\bottomrule
\end{tabular}%
}
\end{table}

The shared exponent does not imply shared performance. \Zippy{} is faster than \Mugatu{} at every simulated size.

\subsection{Torque Scaling}
\label{sec:sub_torque}

\begin{figure}[t]
\centering
\scriptsize
\includegraphics[width=0.91\columnwidth]{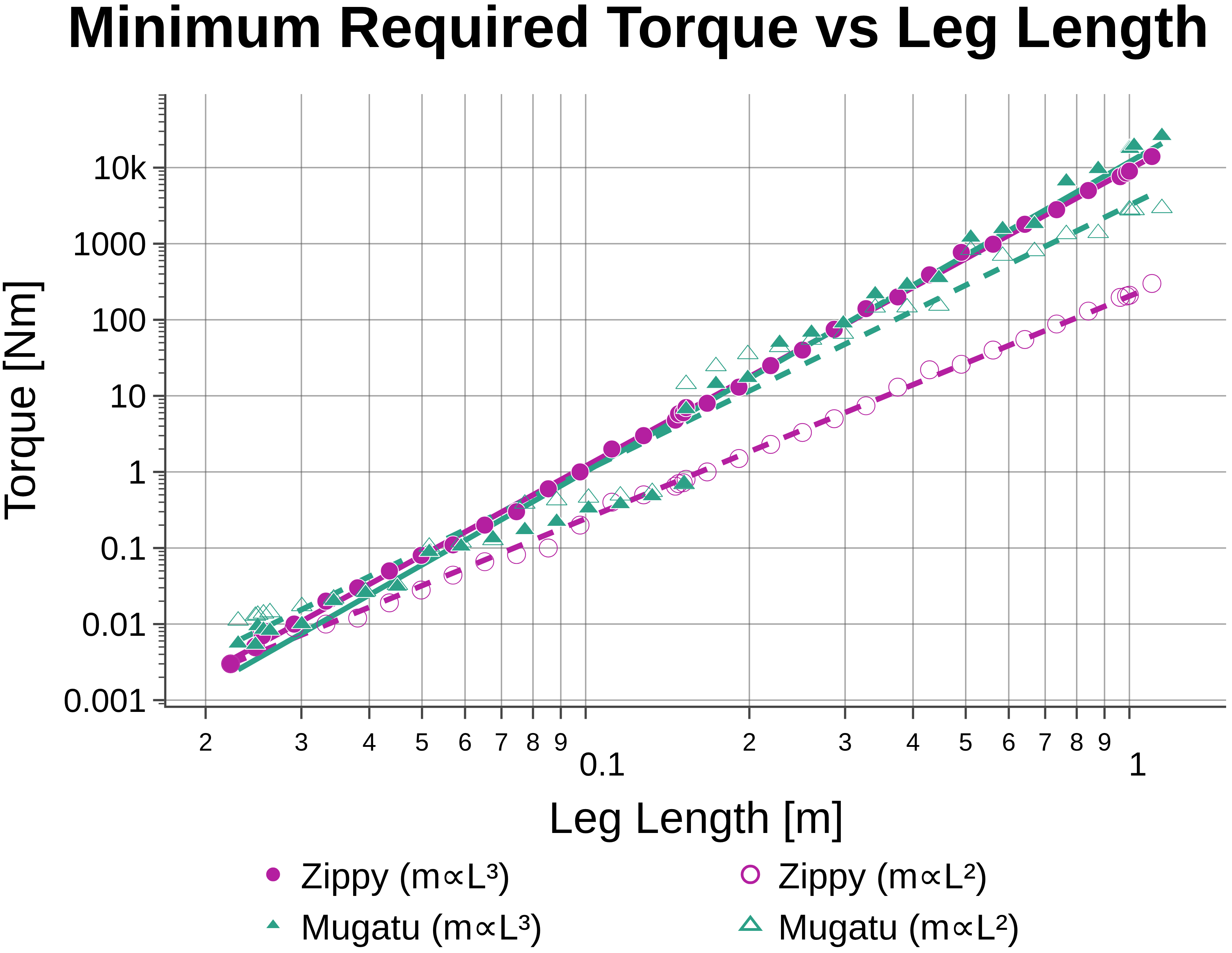}
%\vspace{-2em}
\caption{Simulated minimum required torque of \Zippy{} and \Mugatu{} across scales (see \Cref{tab:simresults}).}
\label{fig:torque}
\end{figure}

Minimum walking torque depends directly on the assumed mass model (\Cref{fig:torque} and \Cref{tab:simresults}). With $m\propto L^2$, the fitted exponents are \num{2.92} for \Zippy{} and \num{3.49} for \Mugatu{}. 
%Budgeting for the worst-case torque for $m \propto L^2$ in \Zippy{} and \Mugatu{}, 
Even in the worst case, we find that the minimum required torque is at the lower end of the torque available in existing motors (\Cref{sec:sub_relatedwork}), suggesting that motors should be able to scale up with the designs. With $m\propto L^3$, the torque exponent increases to \num{3.90} and \num{4.07}, which is at the higher range for available torque. 

Across all cases, \textbf{the common relationship is approximately $\tau\propto mL$}, confirming \cite{miller_clark_IROS2015}, while in no cases do the results support the $\tau \propto mL^2 \propto L^5$ isometric estimate. 
This distinction matters for scaled redesign. A designer who uniformly scales density should budget torque near $L^4$ for a dynamically similar gait; a designer using the mass trend observed among existing bipeds should budget closer to $L^3$.

\subsection{Foot-Shape Scaling and Body Attitude}
\label{sec:sub_footresults}

The foot-shape sweep addresses two questions crucial to passive-dynamic walking: whether a foot permits walking and which foot shape produces the best gait. 
%We find that the boundary of the feasible region scales approximately with leg length, but the forward velocity-maximizing shape and the body motion associated with forward velocity differ between the robots. We also find that retraction, the backward travel during a stride, consistently reduces net progress.

%\subsubsection{Foot Shape Governs Walking Feasibility Across Scales}

%\label{sec:subsub_footfeasibility}

Under either mass scaling, the foot-axis dimensions associated with walking scale as $L^{\num{0.95}}$--$L^{\num{1.03}}$ across all three axes
%, and under $m\propto L^2$ as $L^{\textcolor{black}{\num{0.99}}}$--$L^{\textcolor{black}{\num{1.03}}}$
(\Cref{tab:simresults}). Therefore, \textbf{the feasible region follows geometric similarity, $R \propto L^1$}: a viable foot-shape scaled in proportion to leg length tends to remain viable. The different coefficients for \Zippy{} and \Mugatu{}, however, show that the location of this region is design-specific. \Mugatu{} has similar $X$ and $Y$ axis coefficients, showing the pitching and rolling coupling with passive dynamics. \Zippy{}, in comparison, has higher coefficients for the $X$ axis, showing the necessity of its flatter foot. Note that while feasible foot-shape designs scale proportionally, which design is optimal within that range may be different at different scales (although it is typically within a few percent of the scaled optima). Peak velocity over the foot-shape sweep follows dynamic similarity closely under both mass laws for both robots (\Cref{tab:simresults}).

\begin{figure*}[t]
  \centering
\includegraphics[width=0.78\textwidth]{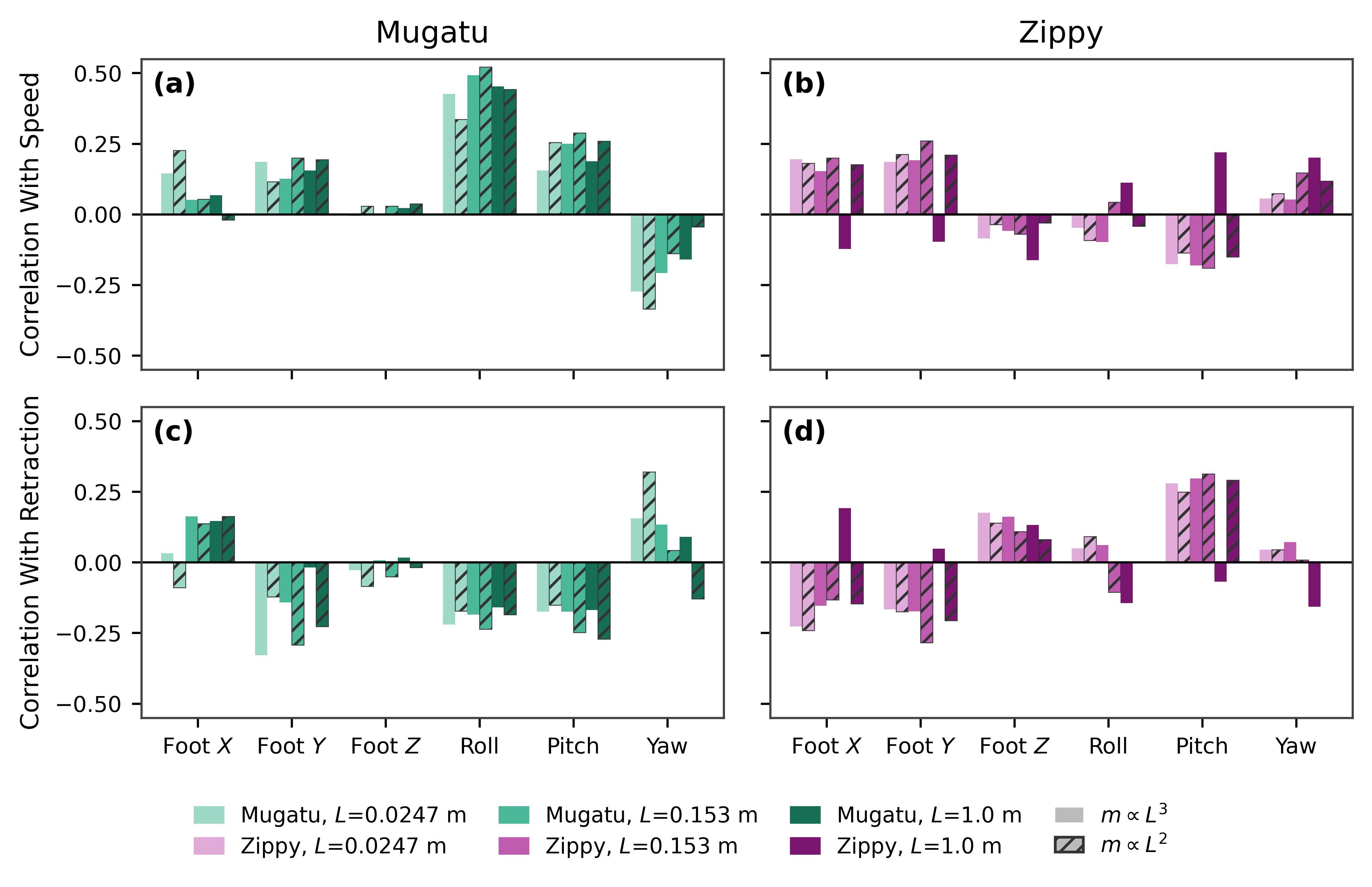}
  \caption{Which foot-axis scales and body dynamics influence forward walking velocity (a,b) and retraction (c,d), shown as correlations at each of the three leg lengths. Solid bars are $m\propto L^3$ and hatched bars $m\propto L^2$. The ranking of drivers is the same under both mass laws.}
  \label{fig:speedretraction}
\end{figure*}

\Cref{fig:speedretraction} shows how foot shape and gait correlate with both speed and retraction fraction, $\rho$. Lateral roll is the strongest speed correlate for \Mugatu{} in every case, showing that body roll is critical to the walking strategy. However, \Zippy{}'s strongest (negative) body-motion correlate is pitch, \textbf{showing that \Zippy{} has a distinct walking behavior from \Mugatu{}}. Foot length and width show similarly small, scale-dependent correlations.

\section{Discussion}
\label{sec:discussion}
% These results are most useful when grouped into three levels of design inference: relationships shared by multiple robots, choices that remain robot-specific, and hypotheses that may transfer to broader classes of legged systems.

\subsection{Shared Scale-Dependent Requirements}
\label{sec:sub_discscale}

Velocity and torque are the clearest shared relationships. Recalling \Cref{eq:pendular_freq}, a geometrically similar walker behaves approximately as a physical pendulum and is mass-independent, giving $v\propto L^{1/2}$ regardless of actuation strategy. This explains both the measured exponent and its insensitivity to the assumed mass model. 

% However, it does not explain why \Zippy{} is faster than \Mugatu{}, suggesting that the coefficient still depends on foot shape, inertia, and control.

The same timescale resolves the torque result. Gravitational torque scales as $\tau = mgL$. Swing inertia gives the same dependence because $\alpha\sim\omega^2\propto L^{-1}$ and therefore
\begin{equation}
\tau_i\sim I\alpha\propto mL^2L^{-1}=mL.
\label{eq:torque}
\end{equation}
Thus, $\tau\propto mL$ is the shared requirement, and $\tau \propto L^3$ or $\tau \propto L^4$ follows only after choosing a mass model. This agrees with the dimensional result of Miller and Clark \cite{miller_clark_IROS2015}. Our survey supplies evidence that $m\propto L^2$ may be a more realistic first estimate for a new biped than constant-density scaling, and therefore $\tau \propto L^3$.

\subsection{Design-Specific Mechanisms for Locomotion}
\label{sec:sub_designmech}

For both \Mugatu{} and \Zippy{}, greater per-stride retraction was consistently associated with lower leg-length-normalized walking speed. Considering all designs, scales, and mass-scaling assumptions, Pearson correlations ranged from $r=\num{-0.69}$ to $\num{-0.86}$ ($n=\num{2018}$-$\num{4143}$). Across foot shape changes, \Mugatu{}'s roll and pitch shape the gait and the resulting body motion increases forward distance traveled and speed. Because of the sinusoidal command and curved foot, increased pitch results in increased yaw. Increased yaw increases retraction, making \Mugatu{} a slower walker (\Cref{tab:simresults} and \Cref{fig:speedretraction}). In contrast, \Zippy{}'s forward walking speed and per-stride retraction are not strongly correlated with roll and yaw, but rather negatively correlated with pitch (See \Cref{fig:speedretraction}). \Zippy{}'s ellipsoidal foot is flatter fore-aft than laterally, and its bang-bang controller with hard stops fixes the forward progress and stride rather than relying on the natural body dynamics like \Mugatu{}.

The reason for this difference in control architecture was a practical consequence of scale: at \Zippy{}'s size, implementing precise closed-loop position control requires packaging sensing and control hardware into a small platform while maintaining the faster actuator response associated with its shorter characteristic timescales \cite{man-zippy-2025}. The added mass and thermal overhead of a motor-encoder or servo assembly are also proportionally more significant at this scale \cite{rodriguez2026}. However, this constraint ultimately proved advantageous: the resulting control strategy produced a different walking mechanism that achieved higher speeds at the given scale.

This explains how two robots can share the same velocity exponent while retaining different absolute speeds and different optimal feet. For \Mugatu{}, foot-shape selection for fast walking should prioritize motion that increases roll and pitch, but is limited to foot shapes that preserve a stable roll-mediated limit cycle; for \Zippy{}, the foot should utilize fore-aft shape to reduce per-stride retraction and cooperate with the stride prescribed by the bang-bang controller.

\subsection{A Scaling Workflow for Robot Redesign}
\label{sec:sub_workflow}

The case study suggests the following procedure for adapting an existing legged robot to a new size:
\begin{enumerate}
    \item Start with isometric scaling as a prior, with all dimensions scaled equally. However, the body dimensions should scale more slowly, as $L_B\propto L^{0.87}$ (\Cref{sec:existingrobots}).
    \item Scale the mass budget using the survey regressions as a prior, e.g.\ $m \propto L^2$ for bipedal designs (\Cref{sec:existingrobots}).
    \item Scale walking velocity as $v \propto L^{1/2}$ and use it to estimate actuation frequency and system speed (\cref{sec:sub_mfv}).
%    \item Geometrically scale contact dimensions to obtain an initial feasible design.
    \item Scale the torque requirement for actuator selection by $\tau \propto mL$ (\Cref{sec:sub_torque}).
    \item Identify available gait mechanisms at the target scale and match the mechanism to the source robot design by measuring body motion, contact motion, and retraction (\Cref{sec:sub_footresults}).
    \item Re-optimize contact morphology around the gait mechanism re-identified at the target scale.
\end{enumerate}

Steps 1-4 are plausible starting points for other gravity-dominated natural legged systems, while steps 5-6 are deliberately system-specific. More highly actuated bipeds, running gaits, compliant legs, and active stabilization may follow different coefficients or even different dominant relationships. 

% The broader contribution of this case study is therefore the separation between transferable physical requirements using natural body dynamics and non-transferable locomotion mechanisms.

\section{Conclusions and Future Work}
\label{sec:conclusions}

This paper began with a design question centered on two robots that claimed to use the same strategy at different scales, but moved differently: why did two geometrically related, single-actuator bipeds require different feet and achieve different relative speeds? The controlled comparison shows that part of the answer is scale-dependent. Both designs follow $v\propto L^{1/2}$, their minimum walking torque follows $\tau\propto mL$, and the boundary of feasible foot-shape scales approximately with leg length. The remaining difference is mechanism-dependent. \Mugatu{} gains speed through lateral rocking, while \Zippy{} relies more directly on its imposed stride and loses speed through pitch and retraction. Geometric scaling can therefore preserve a feasible starting point without preserving the best design.

The survey of existing robots shows that $L_B \propto L^{0.87}$ and $m\propto L^2$ are useful priors for biped design, but the data do not establish a universal robotic law. We conclude from the survey a methodological approach to redesigning legged systems across scales: estimate how mass and gravity-driven requirements scale, then re-identify and optimize the mechanism that produces locomotion at the target size. 

This work provides a starting point for legged robot designs, but can be expanded. First, the controlled results come from two quasi-passive walkers with one hip actuator. More highly actuated systems, non-walking gaits, compliant mechanisms, and multi-legged robots may not share the same dominant dynamics. Second, simulation does not reproduce every scale-dependent hardware effect, including manufacturing tolerances, transmission losses, sensing limits, friction variability, and impact damage. Third, the literature dataset is modest: the field of legged robotics is growing rapidly and it is important to record the capabilities and dimensionalities of these systems for further comparison. In particular, reported speeds between designs involve different gaits and different design objectives; we therefore do not infer a cross-platform velocity law from the survey.

We propose that the next validation step should be a new physical robot designed using the workflow above. A larger version should begin with a component-level $m\propto L^2$ mass target, an $L^{1/2}$ gait period, an $mL$ torque budget, and proportionally scaled feet. Foot-shape and control should then be re-optimized using retraction and body-motion measurements rather than copied from either existing robot. Testing that prediction in hardware would distinguish relationships that genuinely transfer across designs from those that are artifacts of this simulated case study. Extensions to full bipeds and quadrupeds should similarly test the workflow before claiming shared exponents, and should add power and cost of transport to the design requirements.

More broadly, we also argue that progress toward generalizable scaling principles will require greater standardization in how legged robots are described and evaluated. Consistent nomenclature and taxonomies for morphology and locomotion, together with standardized reporting of quantities such as characteristic dimensions, mass distribution, gait frequency, speed, actuator capability, and energetic performance, would make cross-platform comparisons substantially more rigorous. Existing efforts have begun to establish such conventions, but broader adoption would enable larger and more comparable datasets from which scaling relationships can be tested, refined, and eventually translated into reliable design methods. In this sense, developing general scaling laws for robots depends not only on additional experiments, but also on establishing a common language and set of measurements for comparing robotic systems.

%\addtolength{\textheight}{-3cm} 


%%%%%%%%%%%%%%%%%%%%%%%%%%%%%%%%%%%%%%%%%%%%%%%%%%%%%%%%%%%%%%%%%%%%%%%%%%%%%%%%

%%%%%%%%%%%%%%%%%%%%%%%%%%%%%%%%%%%%%%%%%%%%%%%%%%%%%%%%%%%%%%%%%%%%%%%%%%%%%%%%

%%%%%%%%%%%%%%%%%%%%%%%%%%%%%%%%%%%%%%%%%%%%%%%%%%%%%%%%%%%%%%%%%%%%%%%%%%%%%%%%
%\section*{APPENDIX}


\section*{ACKNOWLEDGMENT}
This work includes Claude and ChatGPT use for debugging simulation visualization code as well as generating and debugging figure-plotting scripts.

%%%%%%%%%%%%%%%%%%%%%%%%%%%%%%%%%%%%%%%%%%%%%%%%%%%%%%%%%%%%%%%%%%%%%%%%%%%%%%%%

\bibliographystyle{IEEEtran}
\bibliography{references}

\end{document}